\documentclass[10pt,twocolumn,letterpaper]{article}

\usepackage{3dv}
\usepackage{times}
\usepackage{epsfig}
\usepackage{graphicx}
\usepackage{amsmath}
\usepackage{amssymb}

\usepackage{caption}
\usepackage{multirow}
\usepackage{tabu}
\usepackage[symbol]{footmisc}
\usepackage{enumerate}
\usepackage[inline]{enumitem}

\usepackage[pagebackref=true,breaklinks=true,letterpaper=true,colorlinks,bookmarks=false]{hyperref}

\newcommand\T{\rule{0pt}{2.9ex}}       
\newcommand\B{\rule[-1.2ex]{0pt}{0pt}} 

\threedvfinalcopy 

\def\threedvPaperID{****} 

\ifthreedvfinal\fi
\begin{document}

\title{\textit{To complete or to estimate, that is the question}:\\A Multi-Task Approach to Depth Completion and Monocular Depth Estimation}

\author{Amir Atapour-Abarghouei$^1$ \quad Toby P. Breckon$^{1,2}$\\
	$^{1}$Department of Computer Science -- $^{2}$Department of Engineering\\
	Durham University, UK\\
	{\tt\small \{amir.atapour-abarghouei,toby.breckon\}@durham.ac.uk}}

\maketitle
\thispagestyle{empty}

\begin{abstract}
	Robust three-dimensional scene understanding is now an ever-growing area of research highly relevant in many real-world applications such as autonomous driving and robotic navigation. In this paper, we propose a multi-task learning-based model capable of performing two tasks:- sparse depth completion (i.e. generating complete dense scene depth given a sparse depth image as the input) and monocular depth estimation (i.e. predicting scene depth from a single RGB image) via two sub-networks jointly trained end to end using data randomly sampled from a publicly available corpus of synthetic and real-world images. The first sub-network generates a sparse depth image by learning lower level features from the scene and the second predicts a full dense depth image of the entire scene, leading to a better geometric and contextual understanding of the scene and, as a result, superior performance of the approach. The entire model can be used to infer complete scene depth from a single RGB image or the second network can be used alone to perform depth completion given a sparse depth input. Using adversarial training, a robust objective function, a deep architecture relying on skip connections and a blend of synthetic and real-world training data, our approach is capable of producing superior high quality scene depth. Extensive experimental evaluation demonstrates the efficacy of our approach compared to contemporary state-of-the-art techniques across both problem domains.
\end{abstract}\vspace{-0.5cm}
\section{Introduction}
\label{sec:intro}

With the growing demand for accurate 3D scene understanding as an integral part of various computer vision applications, efficient and accurate depth estimation has received significant attention within the research community in the past few years. Conventional depth estimation techniques such as stereo correspondence \cite{scharstein2002taxonomy}, structure from motion \cite{ding2017fusing}, depth from light diffusion \cite{tao2015depth, woodham1980photometric} and alike have led to significant strides in real-world scene understanding applications. However, pervasive issues and complications ever-present in depth-reliant vision systems (\eg missing depth, temporal and in-scene consistency, intensive computational and calibration requirements and alike), have led to the emergence of entire areas of research focusing on refinement procedures post estimation or capture \cite{abarghouei18review, abarghouei16filling, chodosh2018deep, eldesokey2018propagating, ma2018self, van2019sparse} to render scene depth more useful for downstream applications.
\begin{figure}[t!]
	\centering
	\includegraphics[width=0.99\linewidth]{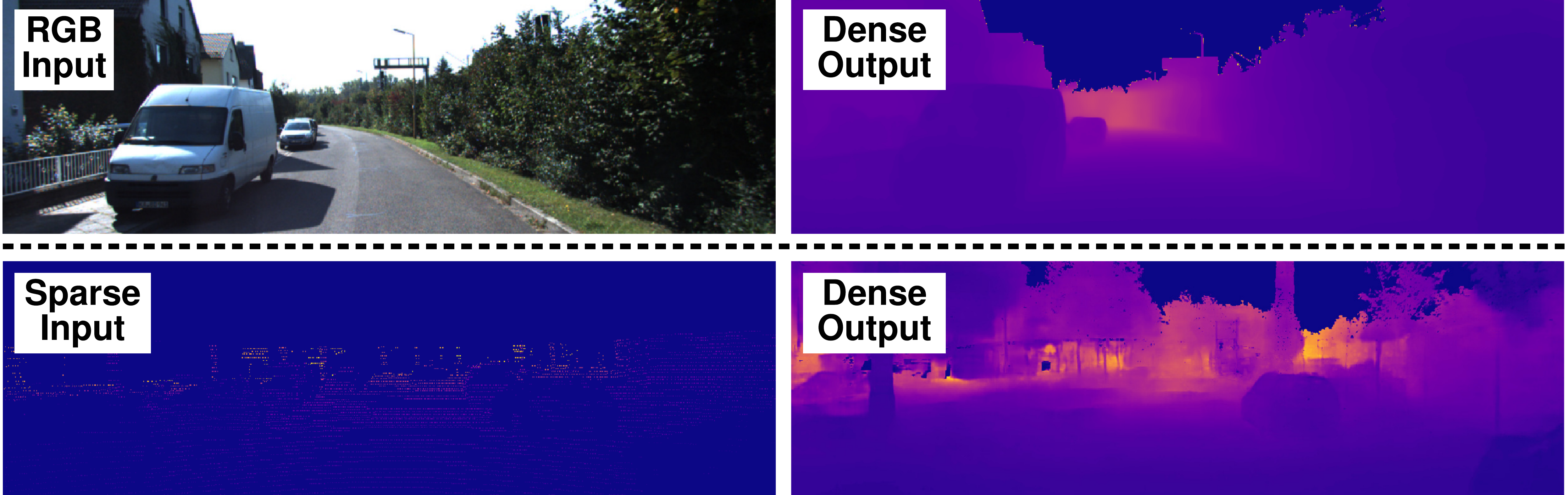}
	\captionsetup[figure]{skip=7pt}
	\captionof{figure}{Exemplar results - a single network architecture facilitates seamless performance of both monocular depth estimation (from RGB, upper) and sparse depth completion (from LiDAR, lower).}%
	\label{fig:taste}\vspace{-0.6cm}
\end{figure}

In recent years, monocular depth estimation (\ie estimating scene depth from a single RGB image) has received widespread attention within both academia and industry as a more effective, economical and innovative alternative to more conventional depth estimation strategies \cite{atapour2018real, abarghouei19segmentwise, eigen2014depth, fu2018deep, monodepth17, kuznietsov2017semi, zhou2017unsupervised}.

In this work, we propose a multi-task depth prediction approach capable of performing sparse depth completion and monocular depth estimation in a joint network trained end to end using a mixture of publicly available synthetic \cite{vkitti} and naturally-sensed real-world \cite{kitti} training data from urban driving scenarios. In order words, this work specifically handles two practical depth estimation/completion scenarios:-
\begin{enumerate*}[label=(\alph*)]
	\item dense depth image estimation from an RGB input (monocular depth estimation) and
	\item sparse to dense depth completion from a sparse LiDAR (laser scanner) input (depth completion).
\end{enumerate*}
Consisting of two sub-networks jointly trained, our approach can seamlessly perform either task without any need for re-training. The first sub-network is solely trained to regress to sparse depth information, similar to that obtained via a 64-channel LiDAR sensor \cite{kitti}. This network carries out its objective based on the information available in the RGB view of the scene, thus mostly focusing on low-level feature extraction to estimate depth values for various objects and components within a constrained region of the scene. This sparse depth output is subsequently utilised by the second sub-network to generate a full dense depth output of the entire scene, requiring high-level inferences and a deeper semantic and geometric understanding of the scene.

During inference in the deployment stage, the entire model can be used as a single unit to perform monocular depth estimation from an RGB colour input, or alternatively, the second sub-network can be utilised alone to yield a dense depth image given a sparse depth input acquired by a LiDAR (laser scanner). Using advances in adversarial training \cite{goodfellow2014generative}, a deep architecture relying on skip connections to preserve high-level spatial features \cite{orhan2017skip, ronneberger2015u} and a combination of synthetic and real-world training data to ensure both the density of the entire depth output and dispensing with any potential domain adaptation requirements \cite{atapour19gan, atapour2018real, zheng2018t2net}, our approach can generate accurate scene depth. In short, our primary contributions are as follows:\vspace{-0.12cm}
\begin{itemize} 
	\setlength\itemsep{1.2mm}
	\item A joint multi-task framework for depth prediction encouraging improved geometric and contextual learning to boost performance.\vspace{-0.22cm}
	\item Monocular depth estimation via adversarial training, a deep architecture with skip connections and a robust compound objective function directly supervised using this framework to outperform prior contemporary work \cite{atapour2018real, atapour2019veritatem, eigen2014depth, monodepth17, kuznietsov2017semi, liu2016learning, zhan2018unsupervised, zhou2017unsupervised}.\vspace{-0.22cm}
	\item Sparse to dense depth completion via the same multi-task model, capable of generating a dense depth output given a sparse depth input captured via a LiDAR sensor with results superior to prior contemporary work \cite{chodosh2018deep, eldesokey2018propagating, mal2018sparse, shivakumar2018deepfuse, uhrig2017sparsity}.\vspace{-0.22cm}
	\item Unique leverage of both synthetic \cite{vkitti} and real-world datasets \cite{uhrig2017sparsity} to ensure high-density complete depth outputs, despite such levels of density not existing in any real-world training images.\vspace{-0.22cm}
	\item Capable of generalising to previously unseen images from different environments since the training data is sampled from varying data domains.\vspace{-0.22cm}
\end{itemize}

\begin{figure*}[t!]
	\centering
	\includegraphics[width=0.99\linewidth]{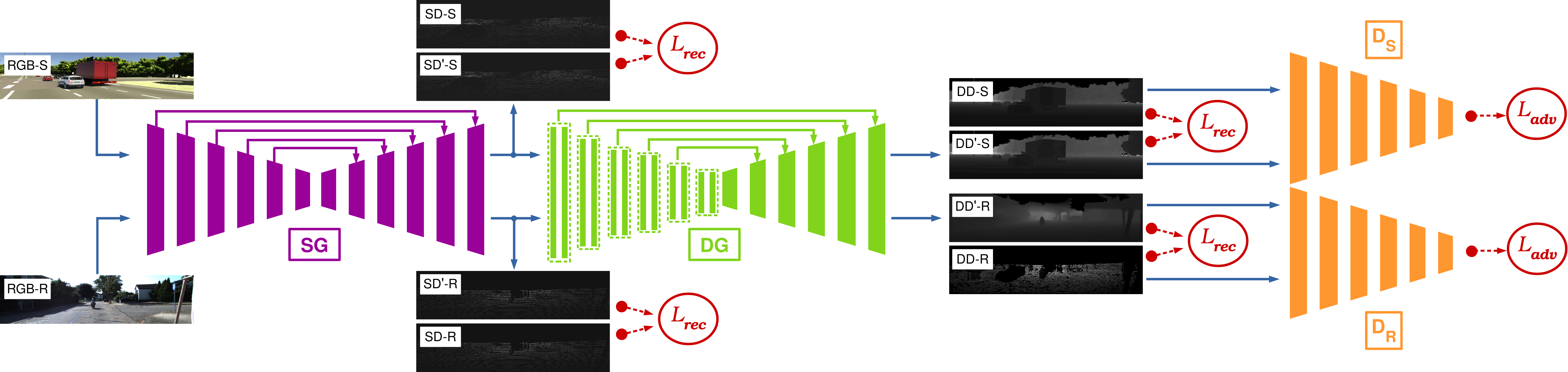}
	\captionsetup[figure]{skip=7pt}
	\captionof{figure}{Overall training procedure of the approach. \textbf{S:} synthetic data from the virtual environment \cite{vkitti}; \textbf{R:} data captured from the real world \cite{uhrig2017sparsity}; \textbf{SD:} sparse depth; \textbf{DD:} dense depth; \textbf{SG:} sparse generator network; \textbf{DG:} dense generator network.}%
	\label{fig:diagram}\vspace{-0.35cm}
\end{figure*}

\section{Related Work}
\label{sec:related}\vspace{-0.010cm}

Here, relevant prior work is considered over two distinct areas, namely monocular depth estimation (Section \ref{ssec:related:MDE}) and sparse depth completion (Section \ref{ssec:related:SDC}).\vspace{-0.05cm}%

\subsection{Monocular Depth Estimation}
\label{ssec:related:MDE}\vspace{-0.05cm}

The emergence of learning-based approaches capable of estimating depth from a single RGB image has caused revolutionary changes in the landscape of 3D scene understanding, leading to significant strides made within the field of monocular depth estimation in recent years. While traditionally, probabilistic graphical models \cite{liu2010single, saxena2006learning, saxena2008make3d} and non-parametric approaches \cite{karsch2014depth, liu2010sift, liu2014discrete} offered promising solutions, their use of hand-crafted features and intensive computational requirements created issues regarding their efficiency and performance capabilities.

With the advent of convolutional neural networks and a growing number of publicly available depth datasets \cite{kitti, silberman2012indoor, song2015sun}, supervised approaches made significant improvements in the state of the art despite prevalent issues in the quality of the ground truth depth for supervision. For instance, \cite{eigen2015predicting, eigen2014depth} generate depth from a two-scale network trained on RGB and depth and \cite{laina2016deeper, li2015depth} offer remarkable performances by providing higher quality outputs.%

On the other hand, more recent techniques began circumventing the need for ground truth depth by reconstructing corresponding views within a stereo correspondence framework to calculate disparity. The work in \cite{xie2016deep3d} generates the right view given the left while producing an intermediary disparity image. Similarly, \cite{monodepth17} employs bilinear sampling \cite{jaderberg2015spatial} and a left/right consistency constraint for improved results. In \cite{zhou2017unsupervised}, depth and pose prediction networks, supervised via view synthesis, are trained to estimate depth and camera motion. The work in \cite{kuznietsov2017semi} produces dense depth by enforcing a model supervised by sparse ground truth depth within a stereo framework via an image alignment loss. Although the training data for the majority of such approaches is abundant and easily obtainable, they still suffer from undesirable artefacts, such as blurring and incoherent content, due to the nature of their secondary supervision.%

More recently, supervised approaches have begun using synthetic training images and are capable of producing better quality depth outputs, despite potential issues with domain shift \cite{atapour2018real, zhao2019geometry, zheng2018t2net}. In this work, we utilise a blend of real world \cite{kitti} and synthetic data \cite{vkitti} in a supervised training approach to accurately estimate depth from a monocular RGB image without the need for any domain adaptation.\vspace{-0.10cm}

\subsection{Sparse Depth Completion}
\label{ssec:related:SDC}\vspace{-0.1cm}

Depth completion can refer to a range of related problems with different input modalities \cite{abarghouei18review}. The existing literature contains a variety of techniques capable of completing relatively dense depth images that contain missing values, such as those utilising exemplar-based depth inpainting \cite{atapour2018extended}, low-rank matrix completion \cite{xue2017depth}, object-aware interpolation \cite{abarghouei17depthcomp}, tensor voting \cite{kulkarni2013depth}, Fourier-based depth filling \cite{abarghouei16filling}, background surface extrapolation \cite{matsuo2015depth}, learning-based approaches using deep networks \cite{atapour19gan, atapour2019veritatem, zhang2018deep}, and alike \cite{chen2014improved, liu2016building}.

However, depth completion can also refer to the problem of generating dense depth information from a scene when only a sparse representation of the scene depth is available. This is of particular interest in robotics applications such as autonomous vehicles where depth sensing technologies such LiDAR are commonly utilised. When depth measurements from such sensors are projected into the camera image space, the available scene depth information accounts for approximately 4\% of the image pixels \cite{uhrig2017sparsity}.

To improve the applicability of such sparse depth measurements, a growing number of novel approaches attempt to estimate dense depth based on the available sparse information. In \cite{uhrig2017sparsity}, sparse convolutions are proposed with input normalisations in mind to take data sparsity into account while training a convolutional neural network. An end-to-end regression model is introduced in \cite{mal2018sparse} to address the problem of sparse depth completion. \cite{eldesokey2018propagating} proposes a constrained convolution operation from which confidence values are propagated through the network. A compressed sensing approach in \cite{chodosh2018deep} utilises a binary mask to filter out unmeasured values in a depth completion framework. The approach in \cite{ma2018self} addresses depth completion by employing a self-supervised training procedure based on sequential RGB and sparse depth images. In \cite{shivakumar2018deepfuse}, a network is proposed that fuses contextual cues learned from RGB and sparse depth inputs to produce dense depth outputs.

\begin{figure*}[t!]
	\centering
	\includegraphics[width=0.99\linewidth]{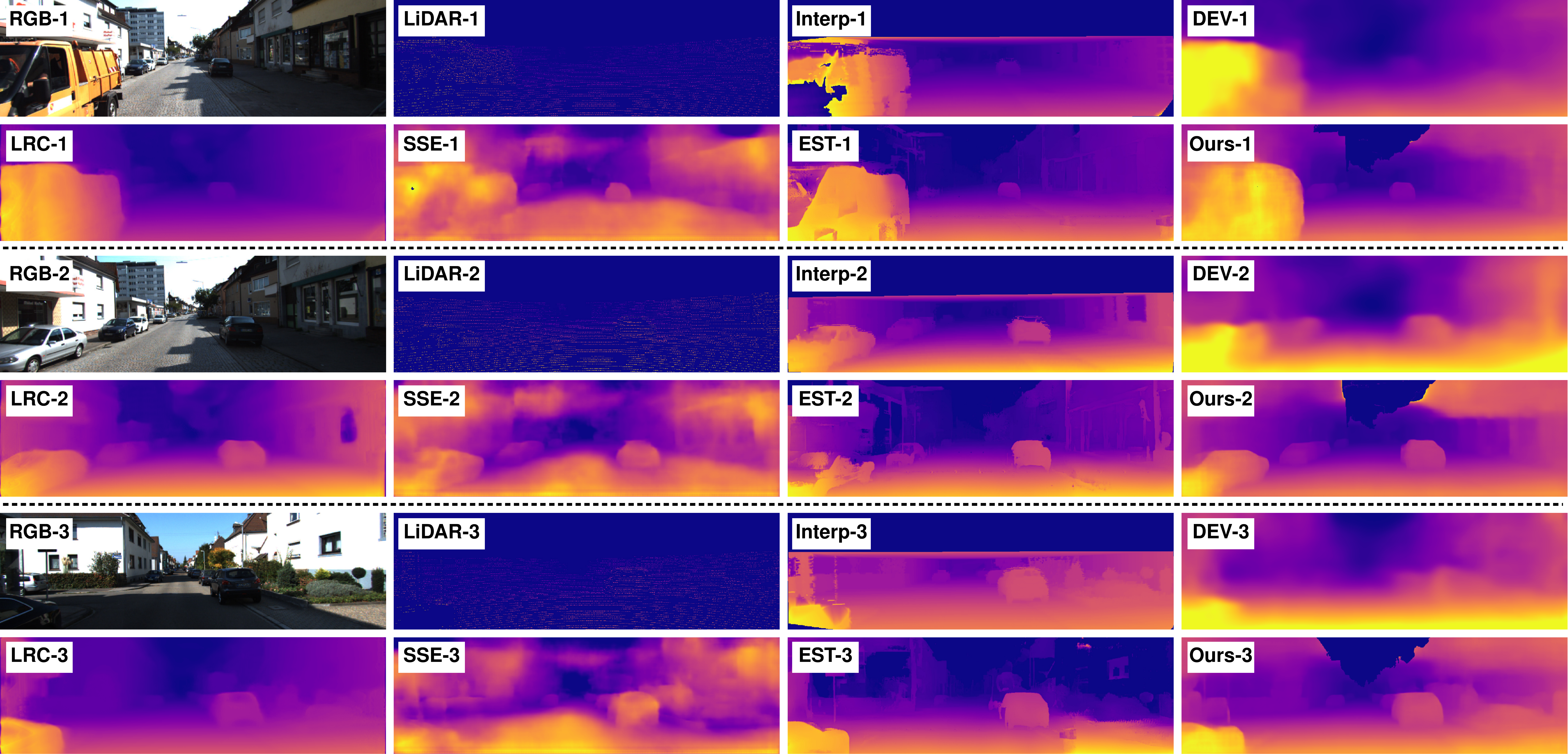}
	\captionsetup[figure]{skip=7pt}
	\captionof{figure}{Comparing the results of our monocular depth estimation approach against \cite{atapour2018real, monodepth17, kuznietsov2017semi, zhou2017unsupervised}. Adjusted disparity images are included for better visibility. \textbf{RGB:} input colour image;  \textbf{DEV:} depth and ego-motion from video \cite{zhou2017unsupervised}; \textbf{LRC:} left-right consistency \cite{monodepth17}; \textbf{SSE:} semi-supervised estimation \cite{kuznietsov2017semi}; \textbf{EST:} estimation via style transfer \cite{atapour2018real}.}%
	\label{fig:estimation}\vspace{-0.3cm}
\end{figure*}

Even though sparse depth completion is not the primary objective of this work, our approach is capable of generating dense depth from a sparse input along with its primary function (monocular depth estimation) and can outperform a variety of prior related work \cite{chodosh2018deep, eldesokey2018propagating, mal2018sparse, shivakumar2018deepfuse, uhrig2017sparsity}.

\section{Proposed Approach}
\label{sec:approach}\vspace{-0.1cm}

The approach proposed here is capable of performing two tasks within a single joint model, monocular depth estimation and sparse depth completion. This has been made possible using two publicly available datasets:- a depth completion dataset based on real-world images \cite{uhrig2017sparsity}, in which relatively dense ground truth depth is created by accumulating measurements made by several laser scans with further consistency enforced between laser scans and stereo reconstruction \cite{hirschmuller2007stereo}; and a synthetic dataset of images captured from a graphically-rendered virtual environment designed for urban driving scenarios \cite{vkitti}.

The primary reason for using synthetic images \cite{vkitti} during training is that despite the increased depth density of the real-world imagery \cite{uhrig2017sparsity}, depth information for the majority of the scene is still missing, leading to undesirable artefacts in regions where depth values are not available. A na\"ive solution would be to only use synthetic data to resolve the issue, but due to differences in the data domains, a model only trained on synthetic data cannot be expected to perform well on real-world images without domain adaptation \cite{atapour2018real, zhang2018deep}. Consequently, we opt for randomly sampling training images from both datasets to force the overall model to capture the underlying distribution of both data domains, and therefore, learn the full dense structure of a synthetic scene while simultaneously modelling the contextual complexity of the naturally-sensed real-world images. 

While the entirety of our approach can be considered a single generative model (G) that predicts full depth, as seen in Figure \ref{fig:diagram}, it is composed of two stages. Each stage relies on a separate sub-network, both trained end to end. Based on the input RGB image, the first network, called the sparse generator (Figure \ref{fig:diagram} - SG), generates a sparse depth image (with non-valid pixels simply set to zero), which the second network, dense generator (Figure \ref{fig:diagram} - DG), subsequently uses to produce the final dense depth output.\vspace{-0.1cm}%

\subsection{Stage 1 - Generating Sparse Depth}
\label{ssec:approach:stage-1}\vspace{-0.1cm}

\begin{figure}[t!]
	\centering
	\includegraphics[width=0.99\linewidth]{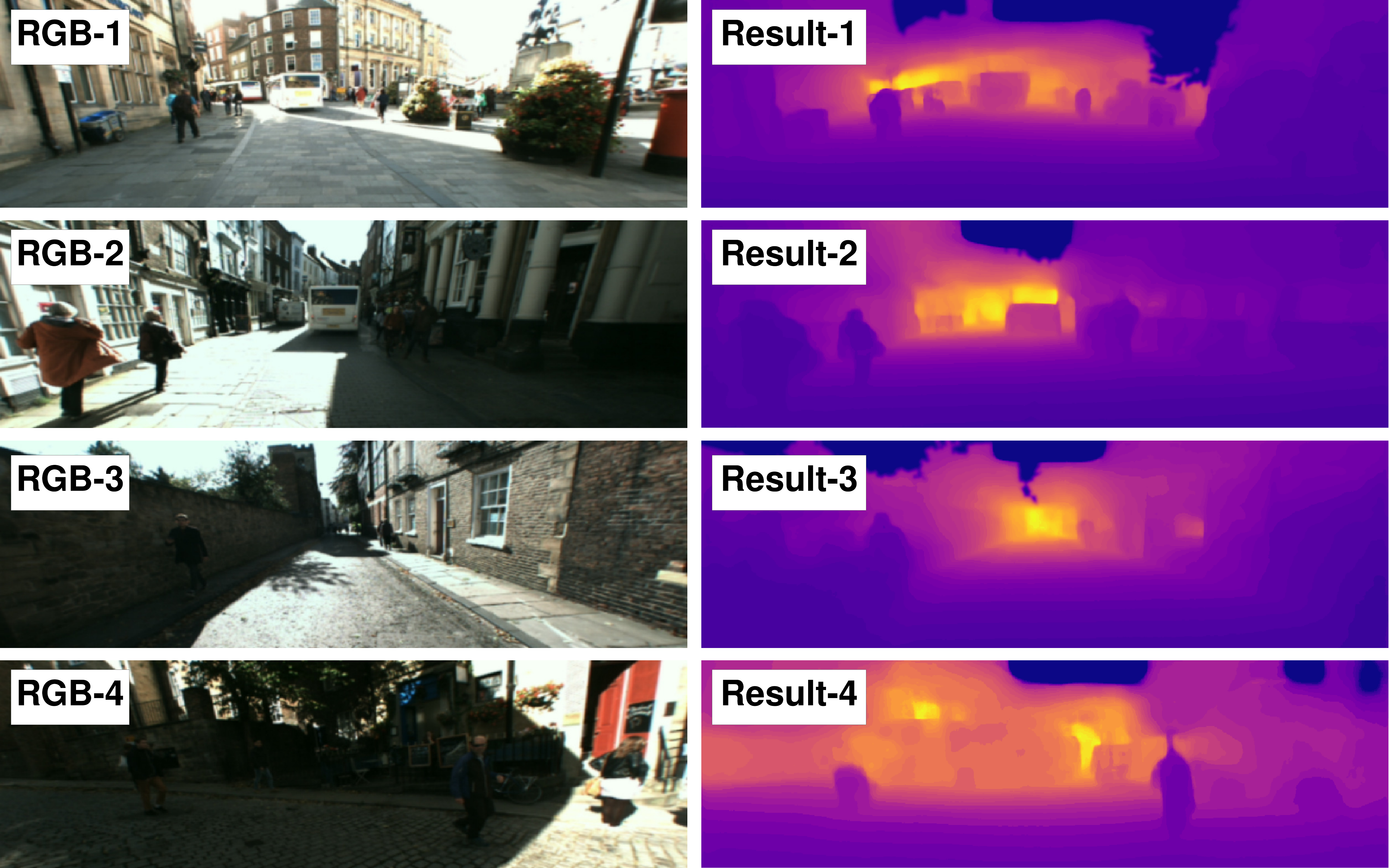}
	\captionsetup[figure]{skip=7pt}
	\captionof{figure}{Demonstrating the generalisation capabilities of the approach using data captured locally from Durham, UK.}%
	\label{fig:durham}\vspace{-0.6cm}
\end{figure}
Our sparse generator (SG) network produces its output by solving an image-to-image translation problem, in which an RGB image is translated to a sparse depth output. More formally, our first generative model ($SG$) encapsulates a mapping function that takes $x$ (RGB image) as its input and outputs $y_s$ (sparse depth) $SG : x \rightarrow y_s$. This can be done by minimising the Euclidean distance between the pixel values of the output ($SG(x)$) and the sparse ground truth ($y_s$). Such a reconstruction objective encourages the model to learn the structural composition of the scene by extracting lower-level features and estimating depth in a constrained window in the scene. This loss is therefore as follows:\vspace{-0.15cm}%
\begin{equation}  
\mathcal{L}_{rec_{SG}}  = ||SG(x) - y_s ||_{1}
\label{eq:loss_rec_sparse}\vspace{-0.4cm}
\end{equation}\\
where $x$ is the input RGB image, $SG(x)$ is the output and $y_s$ the ground truth sparse depth. However, since our training data is randomly sampled from synthetic and real-world images and no sparse ground truth depth is available in the synthetic dataset, it needs to be artificially created. While this could be achieved by training a separate network to predict which pixel values would exist in the sparse depth image (based on the details of the semantic scene objects such as their colour or reflectance qualities), a simpler and just as effective method would be to generate sparse synthetic depth based on a randomly selected sparse depth image from the real-world dataset.

Consequently, before the loss function in Eqn.~\ref{eq:loss_rec_sparse} is calculated for a \emph{synthetic} image, the ground truth sparse depth for said image is generated as follows:
\begin{equation}
	y_{s_{S}}(p)=\begin{cases}
		0, 			& \text{for} \; y_{s_R}(p) = 0		\\
		y_{d_S}(p),	& \text{for} \; y_{s_R}(p) \neq 0
	\end{cases}
	\label{eq:synth_sparse}\vspace{-0.1cm}
\end{equation}\\
where $y_{s_{S}}$, $y_{d_S}$ and $y_{s_R}$ are sparse synthetic depth, dense synthetic depth and sparse real depth images respectively and $p$ denotes the image pixel index. The output of our sparse generator network is subsequently passed to the second sub-network to produce the final result.

\begin{table*}[!t]
	\centering
	\resizebox{\textwidth}{!}{
		{\tabulinesep=0mm
			\begin{tabu}{@{\extracolsep{5pt}}l c c c c c c c c@{}}
				\hline\hline
				\multicolumn{1}{l}{\multirow{2}{*}{Method}} & 
				\multicolumn{1}{c}{\multirow{2}{*}{}} & 
				\multicolumn{4}{c}{Error Metrics (lower, better)} & 
				\multicolumn{3}{c}{Accuracy Metrics (higher, better)} \T\B \\
				\cline{3-6} \cline{7-9}
				&						& Abs. Rel.		& Sq. Rel. 		& RMSE 			& RMSE log 		& $\sigma < 1.25$& $\sigma < 1.25^{2}$& $\sigma < 1.25^{3}$ \T\B \\
				\hline\hline
Train Set Mean 	& \cite{kitti}		   & 0.403 		& 0.530 	& 8.709 	& 0.403 	& 0.593		& 0.776 	& 0.878 \T \\
Eigen \etal		& \cite{eigen2014depth} 		& 0.203 	& 1.548 	& 6.307 	& 0.282 	& 0.702 	& 0.890 		& 0.958 \\
Liu \etal 		& \cite{liu2016learning}		& 0.202		& 1.614 	& 6.523 	& 0.275 	& 0.678 	& 0.895 		& 0.965 \\
Zhou \etal 		& \cite{zhou2017unsupervised}	& 0.208 	& 1.768 	& 6.856 	& 0.283 	& 0.678 	& 0.885 		& 0.957 \\
Atapour \etal 	& 	\cite{atapour2019veritatem}	& 0.193 	& 1.438	   	& 5.887     & 0.234     & 0.836     & 0.930     & 0.958 \\
Godard \etal 	& \cite{monodepth17} 			& 0.148     & 1.344		& 5.927 	& 0.247 	& 0.803 	& 0.922 		& 0.964 \\
Zhan \etal		& \cite{zhan2018unsupervised}	& 0.144     & 1.391		& 5.869     & 0.241		& 0.803		& 0.928		& 0.969 \\
Atapour \etal 	& 	\cite{atapour2018real}	   & 0.110 		& 0.929		& 4.726     & 0.194     & 0.923     & 0.967     & 0.984 \B \\
\hline
Our Approach 	& 		& \textbf{0.080} 		& \textbf{0.836}	& \textbf{4.437}& \textbf{0.157}& \textbf{0.929}& \textbf{0.970}& \textbf{0.985} \T\B \\
				\hline
			\end{tabu}}}
	\captionsetup[table]{skip=7pt}
	\captionof{table}{Numerical comparison of our monocular depth estimation approach over data from \cite{kitti} using the data split in \cite{eigen2014depth}.}
	\label{table:estimation}\vspace{-0.5cm}
\end{table*}

\subsection{Stage 2 - Generating Dense Depth}
\label{ssec:approach:stage-2}\vspace{-0.1cm}

In this stage, our dense generator (DG) network uses the output of the sparse generator $SG(x)$ as its input and generates $y_d$ (dense depth image) $DG : SG(x) \rightarrow y_d$. To ensure that the overall model produces structurally and contextually sound dense depth outputs, a second reconstruction loss component is introduced to ensure the similarity of the final result to the ground truth dense depth:\vspace{-0.15cm}%
\begin{equation}  
\mathcal{L}_{rec_{DG}}  = || DG(SG(x)) - y_d ||_{1}
\label{eq:loss_rec_dense_synthetic}\vspace{-0.6cm}
\end{equation}\\
where $x$ and $y_d$ are the input RGB and dense ground truth depth images respectively. The issue with this loss component arises from the use of synthetic and real-world training data together. While synthetic images are complete and without missing values (except for where necessary, \eg sky and distant objects), real-world ground truth dense depth images from \cite{uhrig2017sparsity} still contain large missing regions. As a result, the reconstruction loss used in Eqn. \ref{eq:loss_rec_dense_synthetic} needs to be reformulated to account for missing values in the real-world ground truth dense depth data. For this purpose, a binary mask, $M$, is created to indicate which pixel values are missing from the ground truth dense depth:\vspace{-0.2cm}
\begin{equation}
	M(p)=\begin{cases}
		0,	& \text{for} \; y_{d_R}(p) = 0		\\
		1,	& \text{for} \; y_{d_R}(p) \neq 0
	\end{cases}
	\label{eq:mask_for_l1}\vspace{-0.2cm}
\end{equation}

Using this binary mask, Eqn. \ref{eq:loss_rec_dense_synthetic} is subsequently reformulated for \emph{real-world} images as follows:\vspace{-0.10cm}
\begin{equation}  
\mathcal{L}_{rec_{DG}}  = || M \odot DG(SG(x)) - y_{d_R} ||_{1}
\label{eq:loss_rec_dense_real}\vspace{-0.5cm}
\end{equation}\\
where $\odot$ is the element-wise product operation. Since Eqn. \ref{eq:loss_rec_dense_synthetic} is used for synthetic images, the model will learn the full structure and context of the scene when the entirety of the scene is available, and at the same time, it will learn to ignore missing regions from real-world images using Eqn. \ref{eq:loss_rec_dense_real}.

With depth prediction being an ill-posed problem (several plausible depth outputs can correctly correspond to an RGB image), exclusively using a reconstruction loss would result in blurry outputs since our overall generative model, $G$, (consisting of both sparse and dense networks) tends to average all possible solutions rather than selecting one, leading to blurring effects. Adversarial training \cite{goodfellow2014generative} can offer a solution to this problem \cite{atapour2018real, dosovitskiy2016generating, isola2016image, yeh2017semantic} since it pushes the model towards selecting single values from the distribution resulting in higher fidelity outputs. Consequently, our overall model ($G$) takes $x$ as its input and outputs fake samples $G(x) = \tilde{y}_d$ while a discriminator ($D$) is adversarially trained to distinguish fake samples $\tilde{y}_d$ from ground truth samples $y_d$. The adversarial loss is thus as follows:\vspace{-0.05cm}%
\begin{equation}
\begin{split}  
\mathcal{L}_{adv} = \min_{G} \max_{D}\ & \mathop{\mathbb{E}}_{x,y_d \sim \mathbb{P}_{d}(x,y_d)} [log D(x,y_d)] + \\
& \mathop{\mathbb{E}}_{x \sim \mathbb{P}_{d}(x)} [log(1 - D(x, G(x)))]
\label{eq:loss_adv}\vspace{-17cm}
\end{split}\vspace{-18cm}
\end{equation}
where $\mathbb{P}_{d}$ is the data distribution defined by $\tilde{y}_d = G(x)$, with $x$ being the generator input and $y_d$ the ground truth. In our approach, we have opted for using two separate discriminators $D_S$ and $D_R$ for synthetic and real-world images respectively (using similar loss as per Eqn. \ref{eq:loss_adv} with the overall adversarial loss being $\mathcal{L}_{adv} = \mathcal{L}_{adv_{S}} + \mathcal{L}_{adv_{R}}$). In our experiments, using a single discriminator for both data types led to stability and convergence issues during training.%
\begin{figure*}[t!]
	\centering
	\includegraphics[width=0.99\linewidth]{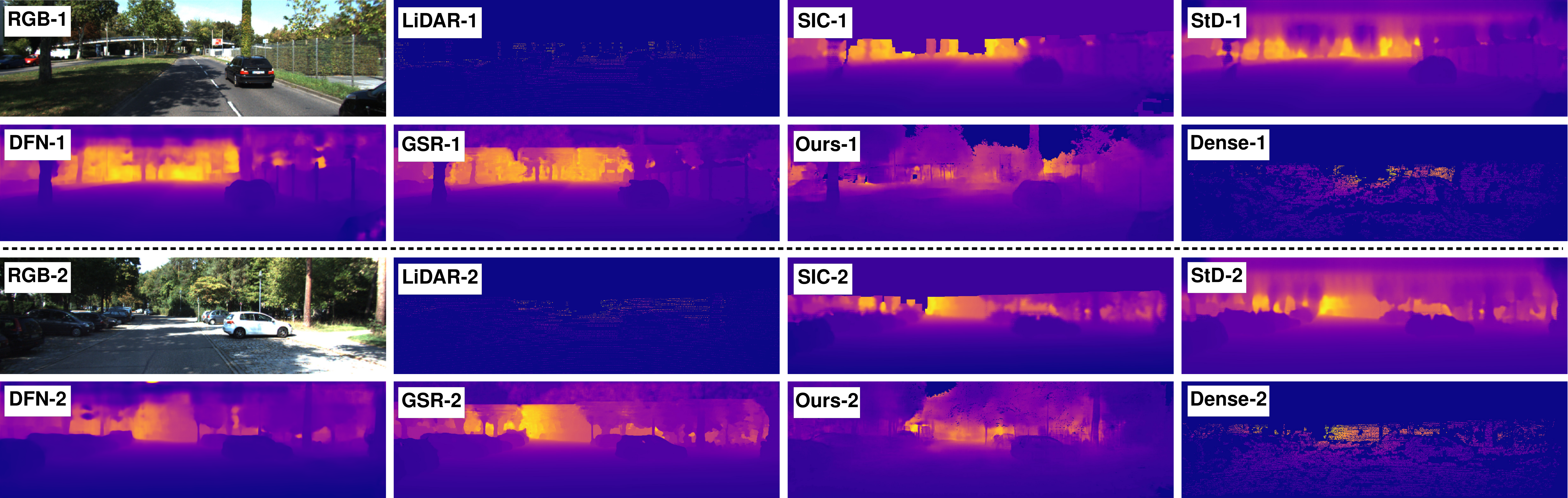}
	\captionsetup[figure]{skip=7pt}
	\captionof{figure}{Comparing our depth completion results against \cite{eldesokey2018confidence, mal2018sparse, uhrig2017sparsity, shivakumar2018deepfuse}. The depth images have been adjusted for better visualization. \textbf{RGB:} input colour image; \textbf{LiDAR:} sparse depth input; \textbf{SIC:} sparsity invariant CNN \cite{uhrig2017sparsity}; \textbf{StD:} sparse to dense completion \cite{mal2018sparse}; \textbf{DFN:} deep fusion network \cite{shivakumar2018deepfuse}; \textbf{GSR:} guided sparse regression \cite{eldesokey2018confidence}; \textbf{Dense:} dense ground truth from \cite{uhrig2017sparsity}.}%
	\label{fig:completion}\vspace{-0.4cm}
\end{figure*}

In addition to this, a smoothing term \cite{monodepth17, heise2013pm} is used to force the model to produce more locally-smooth dense depth results. Depth gradients ($\partial G(x)$) are penalised using $L_1$ regularisation, and an edge-aware weighting term based on input image gradients ($\partial x$) is used to produce smoother depth outputs since image gradients are stronger where depth discontinuities are most likely. The smoothing loss is thus calculated as follows:\vspace{-0.05cm}%
\begin{equation}  
\mathcal{L}_{s}  = |\partial G(x)| e^{|| \partial x ||}
\label{eq:loss_smooth}\vspace{-0.4cm}
\end{equation}\\
where $x$ is the input RGB image and $G(x)$ the output of the overall model. The gradients are summed over vertical and horizontal axes. It is important to note that all the loss components introduced in this section are not only used to train the dense generator (DG) sub-network but the gradients of these loss functions are used to train the entire network end to end, including the sparse generator (SG) sub-network. The overall loss function is therefore as follows:\vspace{-0.01cm}%
{{\small
\begin{equation}  
\mathcal{L} = \lambda_{rec_{SG}}\mathcal{L}_{rec_{SG}} + \lambda_{rec_{DG}}\mathcal{L}_{rec_{DG}} + \lambda_{adv}\mathcal{L}_{adv} + \lambda_{s}\mathcal{L}_{s}
\label{eq:loss_final}\vspace{-0.3cm}
\end{equation}}\\
where the weighting coefficients ($\lambda$) are empirically selected (Section \ref{ssec:approach:implementation}). While the overall model can be used for monocular depth estimation, the second sub-network (DG) can be used alone as a sparse depth completion network since it takes a sparse depth image as its input and can produce an accurate dense depth image as its output.

\subsection{Implementation Details}
\label{ssec:approach:implementation}

Our sparse generator follows an encoder/decoder architecture with every layer containing modules of convolution, BatchNorm and leaky ReLU ($slope=0.2$) with skip connections \cite{ronneberger2015u} between every pair of corresponding layers in the encoder and the decoder (Figure \ref{fig:diagram} - SG). The dense generator follows a somewhat similar architecture, save that in its encoder, residual blocks \cite{he2016deep} form each layer, with the output from each passed to corresponding decoding layers via skip connections. Both discriminators contain an architecture similar to that of \cite{radford2015unsupervised} with each layer including the same modules as those in the generators. All implementation is done in \emph{PyTorch} \cite{pytorch}, with Adam \cite{kingma2014adam} providing the best optimization ($\beta_{1} = 0.5$, $\beta_{2} = 0.999$, $\alpha = 0.0001$). The weighting coefficients in the loss function are empirically chosen, using a basic grid search, to be $\lambda_{rec_{SG}} = 150, \lambda_{rec_{DG}} = 100, \lambda_{adv} = 10, \lambda_{s} = 1$.

\section{Experimental Results}
\label{sec:results}\vspace{-0.1cm}

To rigorously evaluate our approach, we conduct extensive ablation studies and both qualitative and quantitative comparisons with state-of-the-art methods applied to publicly available datasets \cite{kitti, uhrig2017sparsity}. We additionally make use of randomly selected synthetic test images \cite{vkitti} and data captured locally to further evaluate the approach. It is worth noting that using a GeForce GTX 1080 Ti, the entire two passes for monocular depth estimation take an average of 33.4 milliseconds and a single pass through the dense generator for depth completion takes 18.1 milliseconds.%

\begin{table}[!t]
	\centering
	\resizebox{\columnwidth}{!}{
		{\tabulinesep=0mm
			\begin{tabu}{@{\extracolsep{5pt}}l c c c@{}}
				\hline\hline
				\multicolumn{1}{l}{\multirow{2}{*}{Method}} & 
				\multicolumn{1}{c}{\multirow{2}{*}{}} & 
				\multicolumn{2}{c}{Error Metrics (lower, better)} \T\B \\
				\cline{3-4}
				&		& RMSE [$mm$] 			& MAE [$mm$] 	 \T\B \\
				\hline\hline
				Uhrig \etal	      & \cite{uhrig2017sparsity}       & 1729 		   & 503          \T \\
				Chodosh \etal 	   & \cite{chodosh2018deep}         & 1431         & 460             \\
				Eldesokey \etal 	& \cite{eldesokey2018propagating}& 1370         & 377             \\
				Shivakumar \etal  & \cite{shivakumar2018deepfuse}	& 1303 		   & 446             \\
				Eldesokey \etal 	& \cite{eldesokey2018confidence}& 909      & \textbf{210}    \\
				Ma \etal 			& \cite{ma2018self}	            & 879		      & 261             \\
				Van Gansbeke \etal& \cite{van2019sparse}		      & \textbf{802} & 214          \B \\
				\hline
				Our Approach 	& 	                                 & 892          & 243         \T\B \\
				\hline
			\end{tabu}}}
	\captionsetup[table]{skip=7pt}
	\captionof{table}{Comparison of our depth completion approach against \cite{chodosh2018deep, eldesokey2018confidence, eldesokey2018propagating, ma2018self, shivakumar2018deepfuse, uhrig2017sparsity, van2019sparse} using the validation set in \cite{uhrig2017sparsity}. Despite not being the primary focus, our completion approach remains competitive with the state of the art.}
	\label{table:completion}\vspace{-0.6cm}
\end{table}

\subsection{Monocular Depth Estimation}
\label{ssec:results:estimation}\vspace{-0.1cm}

As the primary focus of our proposed approach, our monocular depth estimation model is evaluated against contemporary state-of-the-art approaches \cite{atapour2018real, eigen2014depth, monodepth17, liu2016learning, zhan2018unsupervised, zhou2017unsupervised}. Following the conventions of the existing literature, we use the data split suggested in \cite{eigen2014depth} as the test set.

As seen in Table \ref{table:estimation}, our approach numerically outperforms all comparators across all metrics, mainly due to the superior scene representation learned by the model. It has been established within the literature that de-noising and completion tasks can lead to learning more robust features and a deeper representation of the scene \cite{atapour19gan, pathak2016context, vincent2008extracting, vincent2010stacked}. Since a portion of our model (DG) attempts to complete sparse depth information from the scene, a better understanding of the scene geometric content and semantic context is encapsulated within the model, aiding the approach to not only gain a secondary capability to preform sparse depth completion, but also to perform the primary function of monocular depth estimation more effectively. Qualitative results illustrated in Figure \ref{fig:estimation} also point to the same conclusions. Not only can our approach generate more accurate depth for the entire scene, it does so without undesirable artefacts such as blurring or bleeding effects. As seen in Figure \ref{fig:estimation}, the object boundaries in the results are sharp and crisp, even for more distant scene components.

\begin{figure*}[t!]
	\centering
	\includegraphics[width=0.99\linewidth]{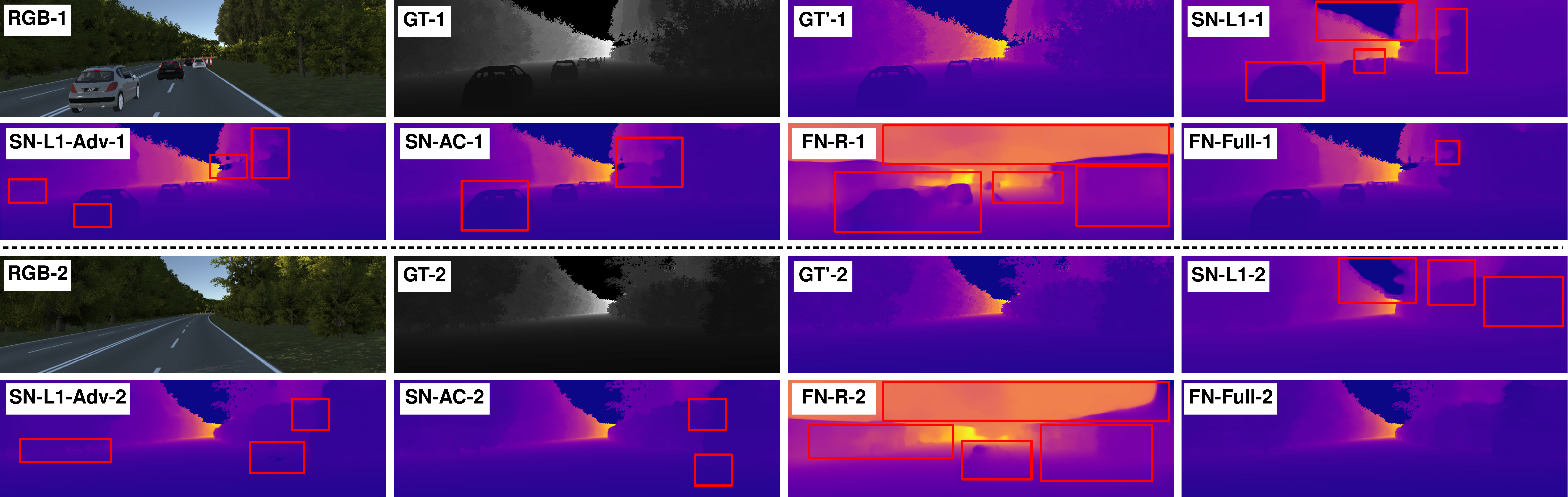}
	\captionsetup[figure]{skip=7pt}
	\captionof{figure}{Comparing the performance of the approach using a randomly selected set of synthetic test images with differing components of the full monocular depth estimation model removed. \textbf{SN:} single network (sparse generator architecture); \textbf{FN:} full network (sparse and dense generators); \textbf{L\textsubscript{1}:} reconstruction loss component; \textbf{Adv:} adversarial loss component; \textbf{AC:} all loss components; \textbf{R:} real-world data only used for training. For clarity, errors in the images are signified by red boxes.
	}%
	\label{fig:ablation}\vspace{-0.4cm}
\end{figure*}

\begin{figure}[t!]
	\centering
	\includegraphics[width=0.99\linewidth]{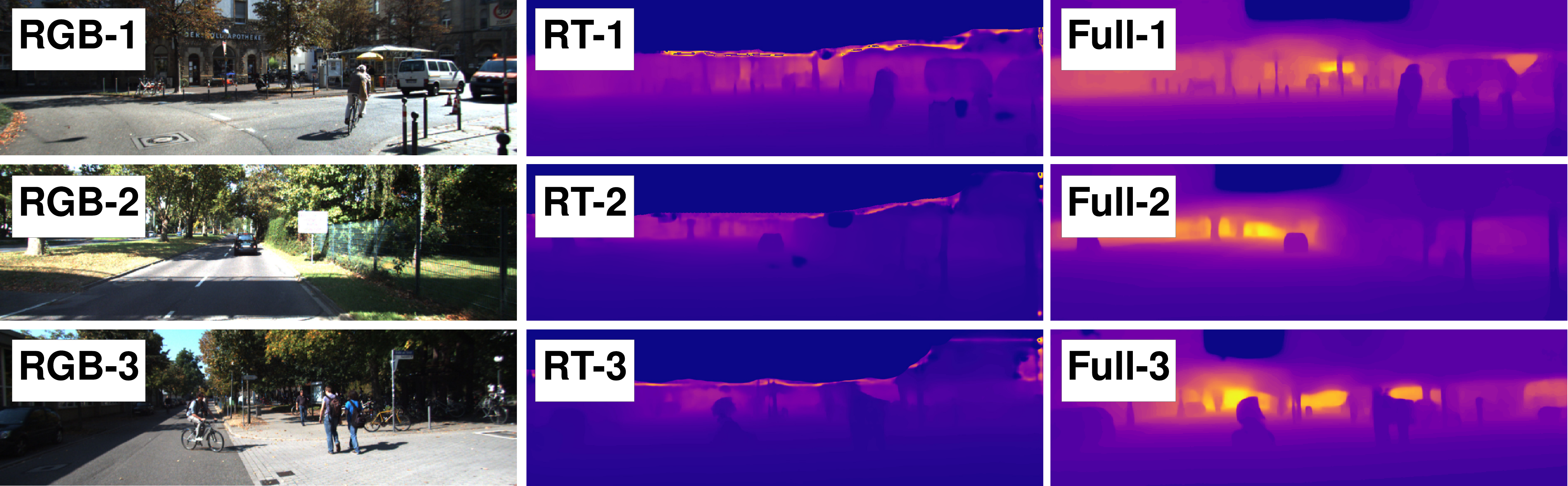}
	\captionsetup[figure]{skip=7pt}
	\captionof{figure}{Demonstrating the importance of using synthetic and real-world data for training using test data from \cite{uhrig2017sparsity}. \textbf{RT:} real-world images used only; \textbf{Full:} full combined dataset of synthetic and real-world data used for training.}%
	\label{fig:real_ablation}\vspace{-0.5cm}
\end{figure}

Additionally, to test the generalisation capabilities of our approach, we apply our model to previously unseen data captured locally in an urban environment in the city of Durham, UK. As seen in Figure \ref{fig:durham}, despite significant differences between the environmental conditions of the training data \cite{vkitti, kitti} and those of the locally captured test data, such as lighting, saturation levels, style, overall shape of the urban environment and alike, the results contain minimal anomalies, are sharp, crisp and very convincing with well-preserved object boundaries and thin structures.

\subsection{Sparse Depth Completion}
\label{ssec:results:completion}\vspace{-0.1cm}

While sparse depth completion is not the primary focus of this work we attempt to extensively evaluate this part of our approach using the publicly available validation dataset in \cite{uhrig2017sparsity} to enable better reproducibility.

As seen in Figure \ref{fig:completion}, due to the use of dense synthetic depth for training and improved scene representation learned by the model, our depth completion approach is capable of predicting depth in the entire scene and visually outperforms all comparators \cite{eldesokey2018confidence, mal2018sparse, shivakumar2018deepfuse, uhrig2017sparsity}. Since the upper regions of the available ground truth images in \cite{uhrig2017sparsity} are missing, all comparators are completely incapable of predicting reasonable depth values for said regions and synthesise erroneous degenerate content. While our approach is certainly not entirely immune to this issue (Figure \ref{fig:limits}), it produces visually improved outputs compared to the other techniques, as seen in Figure \ref{fig:completion}. Numerical results in Table \ref{table:completion} demonstrate that our completion approach quantitatively outperforms many contemporary state-of-the-art completion methods \cite{chodosh2018deep, eldesokey2018propagating, shivakumar2018deepfuse, uhrig2017sparsity} and remains competitive with others \cite{eldesokey2018confidence, ma2018self, van2019sparse}, despite the fact that it is primarily incorporated into our pipeline to improve the main functionality of the approach (monocular depth estimation) and lacks the complex training objectives of many of the comparators.
\begin{table*}[!t]
	\centering
	\resizebox{0.75\linewidth}{!}{
		{\tabulinesep=0mm
			\begin{tabu}{@{\extracolsep{5pt}}l c c c c c c c@{}}
				\hline\hline
				\multicolumn{1}{l}{\multirow{2}{*}{Method}} & 
				\multicolumn{4}{c}{Error (lower, better)} & 
				\multicolumn{3}{c}{Accuracy Metrics (higher, better)} \T\B \\
				\cline{2-5} \cline{6-8}
				&Abs. Rel.		& Sq. Rel. 		& RMSE 			& RMSE log 		& $\sigma < 1.25$& $\sigma < 1.25^{2}$& $\sigma < 1.25^{3}$ \T\B \\
				\hline\hline
				SN/L\textsubscript{1} 	   	& 0.147	& 1.319 & 5.810 & 0.249 & 0.821	& 0.933 & 0.959 \T \\
				SN/L\textsubscript{1}/Adv	& 0.115 & 1.122 & 5.128 & 0.221 & 0.898 & 0.862 & 0.977 \\
				SN/AC 						& 0.108	& 0.982 & 4.911 & 0.198 & 0.913 & 0.962 & 0.980 \B\\
				\hline
				FN/R 						& 0.286 & 1.652	& 6.328 & 0.298 & 0.701 & 0.822 & 0.958 \B\T \\
				\hline
				Full Approach 		 		& \textbf{0.075} & \textbf{0.829}	& \textbf{4.212} & \textbf{0.143} & \textbf{0.951} & \textbf{0.979} & \textbf{0.991} \T\B\\
				\hline
			\end{tabu}}}
	\captionsetup[table]{skip=7pt}
	\captionof{table}{Numerical results with different components of the monocular depth estimation approach. \textbf{SN:} single network (sparse generator architecture); \textbf{FN:} full network (sparse and dense generators); \textbf{L\textsubscript{1}:} reconstruction loss component; \textbf{Adv:} adversarial loss component; \textbf{AC:} all loss components; \textbf{R:} real-world data only used for training.}
	\label{table:ablation}\vspace{-0.4cm}
\end{table*}

\subsection{Ablation Study}
\label{ssec:results:ablation}\vspace{-0.1cm}

To demonstrate the importance of every component of the proposed approach, we re-train our model as varying components of the loss function and the overall approach are removed. It is intuitively expected that using two networks (SG and DG) trained to carry out different stages of a task will lead to better performance than when a single network with half the depth of the architecture is used. We experimentally illustrate this by training a single network with the architecture of the SG to regress to the full dense depth and perform monocular depth estimation in a single pass through the network. Additionally, we remove the components of the loss function to evaluate the influence they have over the performance of the approach. As seen in Table \ref{table:ablation}, the numerical results of experiments over a randomly selected set of synthetic test images (chosen over real images due to their higher level of density) indicate that the model performs better when all elements of the loss function are used during training (SN/AC) and that our full architecture and training procedure (Full Approach) outperforms a single sub-network (SN) by a large margin.

Another important aspect of our approach is incorporating synthetic data into the training process. To evaluate the necessity of this, the full model is trained using real-world data \cite{uhrig2017sparsity} only (FN/R) and the results point to superiority of the joint synthetic/real training data (Table \ref{table:ablation}). Qualitative results in Figure \ref{fig:ablation} also indicate that our full approach with the complete architecture trained on the mixed dataset using the full loss function (Figure \ref{fig:ablation} - FN-Full) outperforms all ablated versions. Since for these experiments, the test images are chosen from the synthetic dataset \cite{vkitti}, but a portion of our ablation studies (FN-R) focuses on the use of real data only, we also evaluated the approach on real-world images \cite{kitti}, and as seen in Figure \ref{fig:real_ablation}, utilising a mixed dataset is more effective than the the use of real images only, even if the test images are selected from the latter.

\section{Discussions and Future Work}
\label{sec:limits}\vspace{-0.1cm}

Even though the use of both synthetic and real-world training data results in better depth predictions, the relatively sparse and flawed real-world ground truth depth images push the underlying model distribution towards degenerate content where depth values are unknown. Examples of such issues can be seen in Figure \ref{fig:limits}, where despite most scene components having correctly been discerned by the model (even the sky), the upper regions still contain incoherent content. Such issues can potentially be addressed in any future work by adding a weighted loss component that can penalise the generator when content is wrongly synthesised in the approximate regions where sky and other distant objects with no depth values are likely found. Additionally, by calculating confidence values for the generated output or propagating confidences through every convolution operation within the network \cite{eldesokey2018confidence, eldesokey2018propagating, kendall2018multi}, this and many other issues such as anomalies and artefacts can be resolved.

\begin{figure}[t!]
	\centering
		\includegraphics[width=0.995\linewidth]{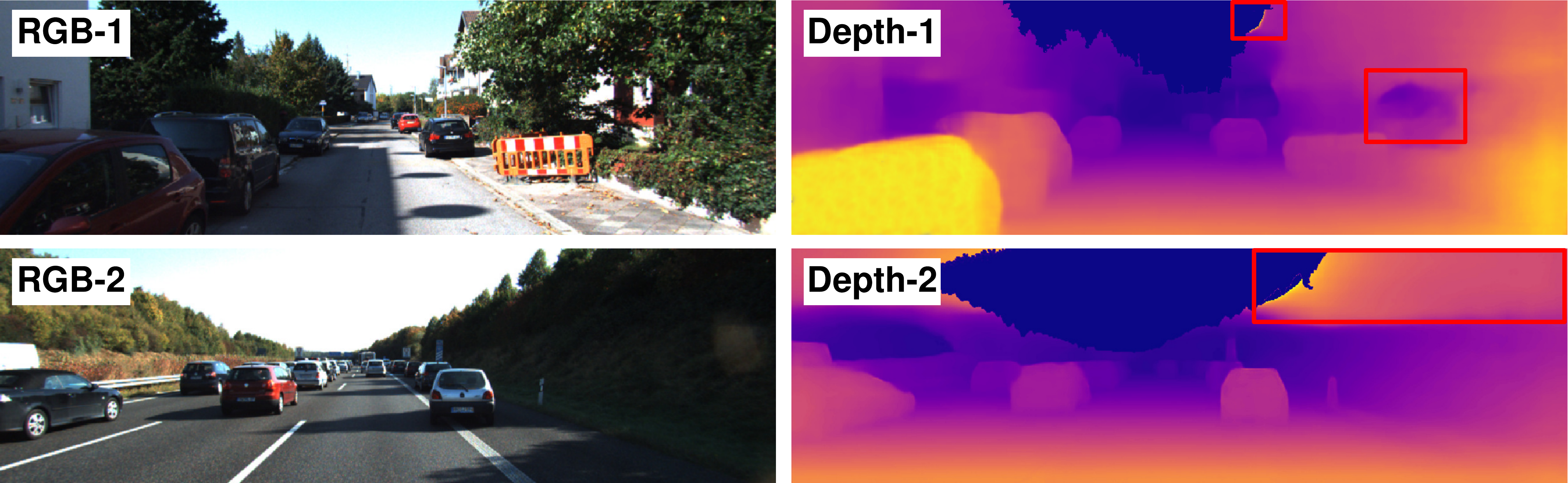}
		\captionsetup[figure]{skip=7pt}
		\captionof{figure}{Examples of the limitations of the approach (red). }%
		\label{fig:limits}\vspace{-0.45cm}
\end{figure}
						
Moreover, while our approach performs both tasks plausibly, it can benefit from improvements in its training procedure. Even though splitting the overall objective of the approach into two stages performed by two sub-networks has led to more robust features within the model and thus its improved results, significant enhancements can be made to the performance by taking advantage of the abundance of information available within the training data \cite{vkitti, kitti}. Inspired by \cite{ma2018self, zhou2017unsupervised} and using the sequential order of frames available in \cite{vkitti, kitti}, photometric transformations and temporal continuity can provide highly beneficial supervisory signals to enforce a deeper contextual learning of the scene.

\section{Conclusion}
\label{sec:conclusion}\vspace{-0.1cm}

Here, we propose a multi-task model that can perform two fundamental scene understanding tasks:- sparse depth completion and monocular depth estimation. This is accomplished using two sub-networks jointly trained on a mixture of publicly available synthetic \cite{vkitti} and natural real-world \cite{uhrig2017sparsity} training data from urban driving scenarios. The first network within the overall pipeline attempts to regress to a sparse depth image, not unlike those generated by projecting depth measurements captured via a LiDAR sensor into image space. This sparse depth output produced by the first sub-network is subsequently passed into the second sub-network which generates a full dense depth image of the entire scene.

The low-level feature extraction and high-level inferences carried out by these two networks lead to better representation learning within the model and consequently its superior performance. Additionally, the entire model can be used to perform monocular depth estimation or, alternatively, the second sub-network can be utilised alone to carry out sparse depth completion. Using adversarial training, a deep architecture with skip connections and a blend of synthetic and real-world training data to guarantee the accuracy and density of the depth output, our approach can produce high quality scene depth. Our extensive experimental evaluation demonstrates the efficacy of our approach compared to contemporary state-of-the-art methods across both domains of monocular depth estimation \cite{atapour2018real, atapour2019veritatem, eigen2014depth, monodepth17, kuznietsov2017semi, liu2016learning, zhan2018unsupervised, zhou2017unsupervised} and sparse depth completion \cite{chodosh2018deep, eldesokey2018propagating, mal2018sparse, shivakumar2018deepfuse, uhrig2017sparsity}.

\textit{We kindly invite the readers to refer to the \href{https://vimeo.com/351624727}{\textbf{video}}: \href{https://vimeo.com/351624727}{https://vimeo.com/351624727} for more information and larger improved-quality result images and video sequences.}


{\small
\bibliographystyle{ieee}
\bibliography{refs}
}

\end{document}